\documentclass[10pt,journal,compsoc]{IEEEtran}
\ifCLASSOPTIONcompsoc
  \usepackage[nocompress]{cite}
\else
  \usepackage{cite}
\fi
\usepackage[figuresright]{rotating}
\usepackage{multirow}
\usepackage{graphicx}
\usepackage{subfigure}
\usepackage{lineno}
\usepackage{color}
\usepackage{hhline}
\usepackage{colortbl}
\usepackage{xcolor}
\usepackage{array}
\usepackage{tabularray}
\usepackage{booktabs}
\usepackage{amssymb}
\usepackage{algorithm} 
\usepackage{algorithmic} 
\usepackage{amsmath}
\usepackage{ragged2e}
\usepackage{tcolorbox}
\usepackage{booktabs}

\ifCLASSINFOpdf
\else
\fi
\begin{document}

\title{Automatic Counterfactual Augmentation for Robust Text Classification Based on Word-Group Search}
%
%
%
%

\author{Rui Song,
		Fausto Giunchiglia,
        Yingji Li,
        and Hao Xu*
\IEEEcompsocitemizethanks{\IEEEcompsocthanksitem Rui Song is with School of Artificial Intelligence, Jilin University. Yingji Li and Hao Xu are with College of Computer Science and Technology, Jilin University, Changchun, China. E-mail: \{songrui20, yingji21\}@mails.jlu.edu.cn, xuhao@jlu.edu.cn\protect\\
Fausto Giunchiglia is with Department of Information Engineering and Computer Science, University of Trento, Italy. E-mail: fausto@disi.unitn.it}
}

%
%

\markboth{Journal of \LaTeX\ Class Files,~Vol.~14, No.~8, August~2015}%
{Shell \MakeLowercase{\textit{et al.}}: Bare Demo of IEEEtran.cls for Computer Society Journals}
%



\IEEEtitleabstractindextext{%
\justify
\begin{abstract}
Despite large-scale pre-trained language models have achieved striking results for text classificaion, recent work has raised concerns about the challenge of shortcut learning. In general, a keyword is regarded as a shortcut if it creates a superficial association with the label, resulting in a false prediction. Conversely, shortcut learning can be mitigated if the model relies on robust causal features that help produce sound predictions. To this end, many studies have explored post-hoc interpretable methods to mine shortcuts and causal features for robustness and generalization. However, most existing methods focus only on single word in a sentence and lack consideration of word-group, leading to wrong causal features. To solve this problem, we propose a new Word-Group mining approach, which captures the causal effect of any keyword combination and orders the combinations that most affect the prediction. Our approach bases on effective post-hoc analysis and beam search, which ensures the mining effect and reduces the complexity. Then, we build a counterfactual augmentation method based on the multiple word-groups, and use an adaptive voting mechanism to learn the influence of different augmentated samples on the prediction results, so as to force the model to pay attention to effective causal features. We demonstrate the effectiveness of the proposed method by several tasks on 8 affective review datasets and 4 toxic language datasets, including cross-domain text classificaion, text attack and gender fairness test. 
\end{abstract}

}

\maketitle

\IEEEdisplaynontitleabstractindextext

%
\IEEEpeerreviewmaketitle

\section{Introduction}\label{sec:introduction}

\IEEEPARstart{T}{ext} classification is a basic natural language processing (NLP) task which has been widely used in many fields, such as sentiment classification~\cite{liu2014tasc}, opinion extraction~\cite{zhou2016cminer}, rumor detection\cite{song2019ced}, and toxic detection~\cite{vaidya2020empirical}. Recent studies have shown that fine-tuning of large-scale pre-training language models (LPLMs) can achieve optimal text classification results, such as BERT~\cite{DevlinCLT19}, ALBERT~\cite{LanCGGSS20}, and RoBERTa~\cite{RoBERTa}. However, some work has raised concerns that existing text classification models often suffer from spurious correlations~\cite{WangC20, WangSY022}, or called shortcut learning~\cite{DuMJDDGSH21}. Although usually without compromising the prediction accuracy, shortcut learning results in low generalization of out-of-distribution (OOD) samples and low adversarial robustness~\cite{ZellersBSC18}.

Consider a widely used example "\textit{This Spielberg film was wonderful}", the term \textit{Spielberg} may be a shortcut, since it often appears alongside positive comments, even though it is not a reliable causal feature that causes the results~\cite{WangC20}. This shortcut fails once the model is migrated to unfriendly dataset to \textit{Spielberg}. A more worth noting example comes from the scenario of toxic text detection. Here, "\textit{They are good at making money}" is not regarded as a toxic description, but by replacing \textit{They} with \textit{Jews}, the example may be seen as toxic~\cite{WiegandRE21}. The excessive focus on words related to certain groups leads to stereotypes, which makes unfairness to the relevant groups. Therefore, more and more studies work on shortcut mitigation and robustness improvement. 

In recent years, it has been proved that counterfactual augmentation can effectively improve the robustness of the classifier to shortcuts~\cite{LuMWAD20, WangC21}. Models trained on augmented data appear to rely less on semantically unrelated words and generalize better outside the domain~\cite{KaushikSHL21}. Therefore, the human-in-the-loop process is designed to take advantage of human knowledge to modify text and obtain opposite labels for counterfactual augmentation~\cite{KaushikSHL21}. But due to the high cost of human labor, many methods of automatic counterfactual augmentation have also been developed~\cite{WangC21, Yadav0GG22, ChoiJHH22}. Edits against auto-mined causal features are used to obtain counterfactual samples. 

\begin{figure}[t!]
	\begin{center}
		\includegraphics[width=0.45\textwidth]{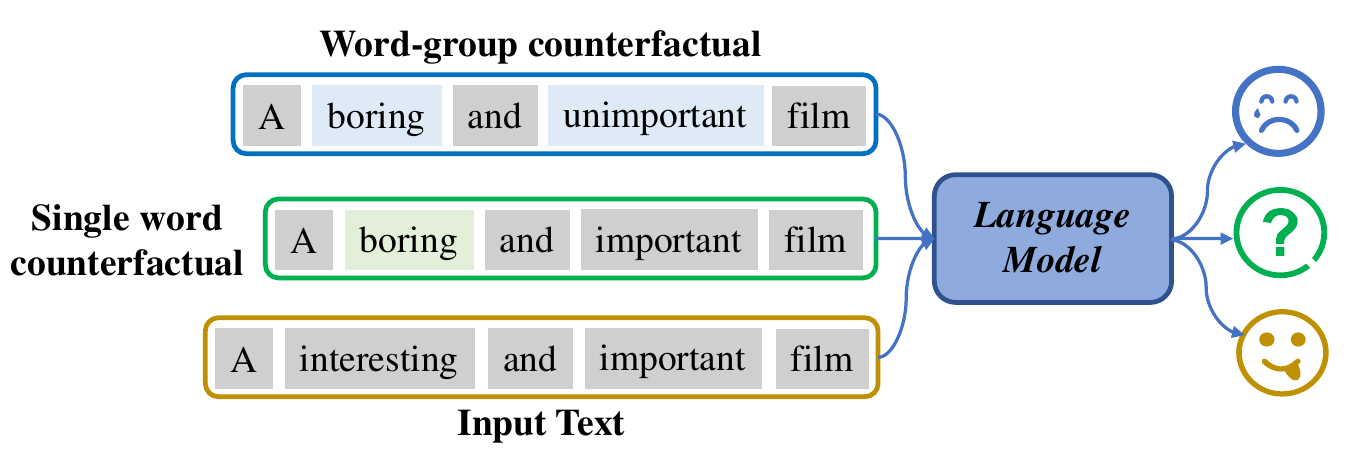}
		\caption{An example to illustrate the effect of word-group on prediction results. It is important to emphasize that the word-group here is a combination of any tokens. It is not mandatory for tokens to be adjacent. }
		\label{fig:example}
	\end{center}
\end{figure} 

\textbf{However, the existing approaches still face two problems. Firstly, they overconsider the contribution of single token and ignore the influence of word-groups. Second, automatically generated counterfactual samples may not have true opposite labels, which can also negatively affect model robustness.} As a result, the automatic counterfactual samples may not be insufficiently flipped due to the omission of causal features, further affecting the true semantics of the counterfactual samples. As shown in Figure~\ref{fig:example}, the emotional slant of a film criticism is determined by both \textit{interesting} and \textit{important}, but a counterfactual for a single word simply replaces \textit{interesting} with \textit{boring}, which results in a sentence with contradictory semantics, thus misleading the model. While a sensible automated counterfactual framework should be able to find the corresponding word-group to generate a true semantic flip sample. 

Based on the above observation, a causal word-group mining method is proposed in the paper, which purpose is to search for the set of keywords that have the most impact on the prediction. In order to avoid some insignificant words from negatively affecting the search efficiency, a gradient-based post-hoc analysis method~\cite{SikdarBH20} is adopted to obtain the candidate causal words of the current sample. Subsequently, a beam search method based on candidate causal words is proposed, whose goal is to counterfactual flip a word group to maximize the change in the probability distribution of predicted logits. This change on the predicted logits is known as \textbf{Causal Effect}. The limited search width and depth ensure the mining efficiency of word-group. 

Moreover, we propose an \textbf{A}utomatic \textbf{C}ounterfactual enhanced multi-instance contrastive learning framework based on \textbf{W}ord-\textbf{G}roup (\textbf{ACWG}). Specifically, for each sample, automatic counterfactual augmentation is performed on the searched word-groups to obtain enhanced samples that are semantically opposite to the original sample. While random masking of some non-causal candidates allows a semantically identical positive sample. Based on the above augmented results, a multi-instance contrastive learning framework is proposed to force language models to rethink semantically identical and opposite samples. To mitigate potential errors from a single word-group augmentation, we select the top $k$ word-groups with the largest causal effect, and jointly optimize the loss of comparative learning through an adaptive voting mechanism. To verify the generalization and robustness of the proposed method, cross-domain text classification and text attack experiments are performed on 12 public datasets, including 8 sentiment classification datasets and 4 toxic language detection datasets. In summary, the contributions of this paper are as follows:
\begin{itemize}
	\item We propose a word-group mining method to overcome the disadvantage of existing robust text classification methods based on automatic causal mining which only focus on the causal feature of a single keyword.
	\item Based on the word-group mining, we further propose an automatic counterfactual data augmentation method to obtain the opposite semantic samples by counterfactual substitution of the word-groups.
	\item Furthermore, we propose a word-groups-based contrastive learning method, which aims to extract stable decision results from multiple word-groups by using a automatic voting mechanism.
	\item Experimental results on 12 public datasets and 3 common used large-scale pre-training language models confirm the validity of the proposed method.
\end{itemize}

\section{Related Work}\label{sec:work}
We introduce some of the work related to the proposed methods in this section, including identification of shortcuts and causal features, and approaches to use them to improve model robustness. 

\subsection{Shortcuts and Causal Features}
How to identify shortcuts and causal features in text is the premise of many robust text classification approaches. One of the most intuitive ways is to use human prior knowledge to label keywords or spurious correlation patterns~\cite{ZhangMW16, Zhao2021, RibeiroWG021}. There are also approaches to make better use of human prior knowledge by designing human-in-the-loop frameworks~\cite{LuYMZ22}. But these methods rely on manual labor and have poor scalability. Therefore, interpretable methods are adopted to facilitate automatic identification of robust/non-robust region at scale, e.g. attention score~\cite{MoonMLLS21}, mutual information~\cite{DuMJDDGSH21} and integrated gradient~\cite{TeneyAH20, RobinsonSYBJS21}. Besides, counterfactual causal inference is also used to determine the importance of a token by adding perturbation to the token~\cite{GargR20, RobinsonSYBJS21}. If the perturbation of a token has a greater impact, the higher the contribution of the token to the prediction result. Some work also seeks to obtain more explicit shortcuts by further integrating various interpretable methods~\cite{DuMJDDGSH21, WangSY022}. 

\subsection{Shortcut Mitigation and Robust Model Learning}
Multiple approaches have been studied for shortcut mitigation and robust model learning such as domain adaptation~\cite{LiMZLLW23} and multi-task learning~\cite{uLGH20}. Under the premise of given shortcuts or causal features, then it is easy to guide the model correctly by adversarial training~\cite{chai2023additive}, reweighting~\cite{nam2020learning}, Product-of-Expert~\cite{Sanh0BR21}, knowledge distillation~\cite{UtamaMG20}, keyowords regularization~\cite{MoonMLLS21} and contrastive learning~\cite{RobinsonSYBJS21}. Recently, researchers have developed counterfactual data augmentation methods to build robust classifiers, achieving state-of-the-art results~\cite{LuMWAD20}. 

Similarly, counterfactual augmentation can be divided into manual and automatic parts. The former relies on human prior knowledge. \cite{GargPLTCB19} counterfactually augments the sample with predefined tokens to improve the fairness of the model. \cite{KaushikHL20} builds a human-in-the-loop system by crowd-sourcing methods to counterfactually augment samples, while improving the robustness and extraterritorial generalization ability of the model. The latter automatically looks for causal features in the sample and flips them to generate counterfactual samples. \cite{NgCG20} generates synthetic training data by randomly moving a pair of corruption and reconstruction functions over a data manifold. \cite{GargR20} uses a masked language model to perturb tokens to obtain adversarial examples. \cite{WangC20, WangC21} obtain counterfactual data by substituting antonyms for words that are highly correlated with the predicted results. Counterfactual texts are assigned to opposite labels and helps train a more robust classifier. \cite{Yadav0GG22} learns the rules by logical reasoning and gives faithful counterfactual predictions. C2L make
a collective decision based on a set of counterfactuals to overcome shortcut learning\cite{ChoiJHH22}. AutoCAD guides controllable generative models to automatically generate counterfactual data~\cite{WenZZ0H22}. 

Similar to previous work, our approach is based on valid interpretable analysis. But the difference is that we automatically generate counterfacts by searching for word groups with the greatest causal effect, rather than just focusing on the effects of individual words. Then, multiple word-groups vote adaptively to learn the impact on the model, reducing the potential for miscalculation from a single word-group. 

\section{Method}\label{sec:method}
\begin{figure*}[t!]
	\begin{center}
		\includegraphics[width=0.95\textwidth]{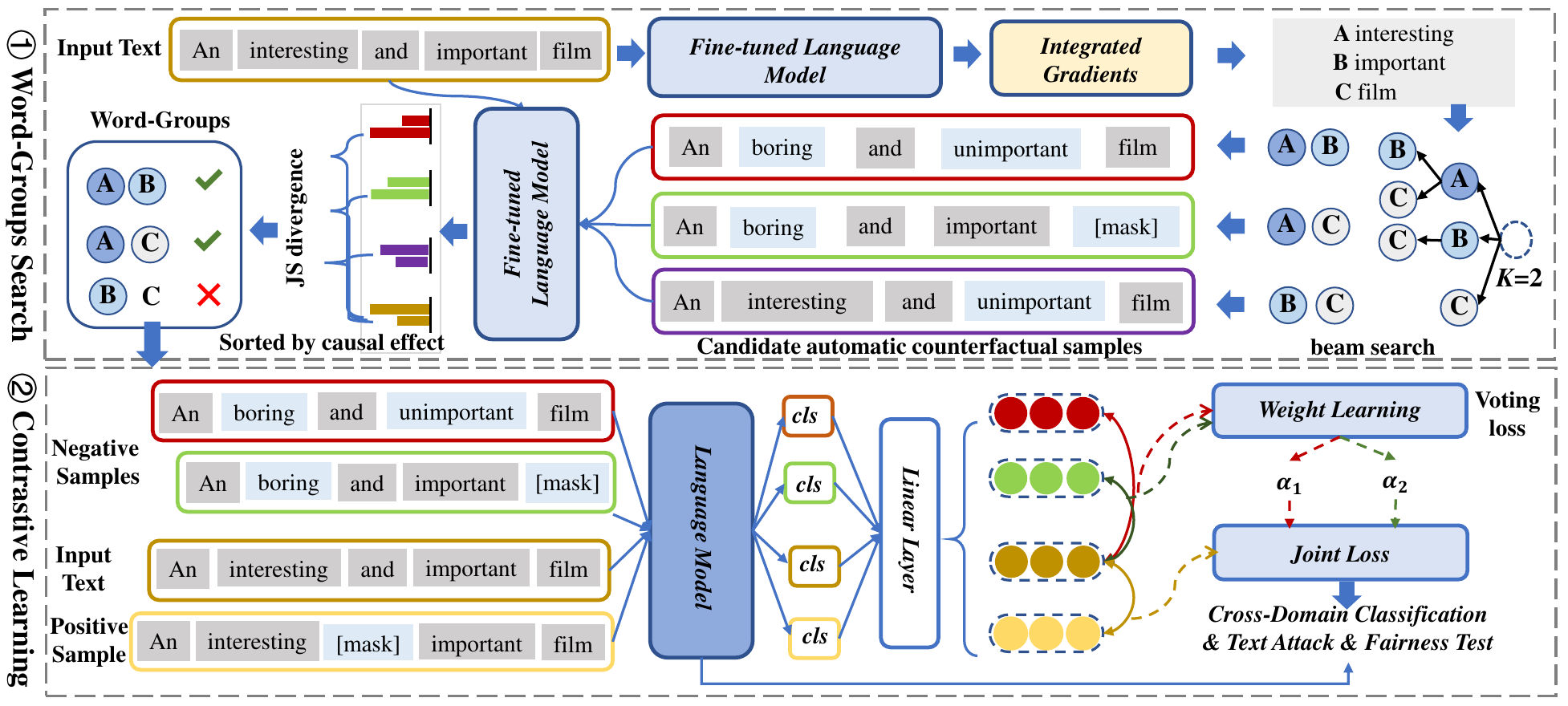}
		\caption{The overall framework of the proposed ACWG. To simplify, we replace the candidate causal words \textit{interesting}, \textit{important}, and \textit{film} with \textbf{A}, \textbf{B} and \textbf{C}, respectively. The same samples, representations, and their logits are represented in the same color. }
		\label{fig:model}
	\end{center}
\end{figure*} 
In this section, we first define the model-related symbols in detail. Then, the detailed framework of the model is introduced in Figure~\ref{fig:model}. The overall framework of ACWG is divided into two parts. First, word-groups search is performed by maximum the causal effect of the language model. Subsequently, a contrastive learning with multiple samples is performed through the searched word-group to learn robust sample representations. 

\subsection{Task Definitions}\label{sec:def}
In this paper, we focus on cross-domain text classification, which aims to fine-tune the language model $\mathcal{M}$ on the training set of the source domain $\mathcal{X}^{train}_{source}$, and produce a trained model $\tilde{\mathcal{M}}$ and a mapping function $f_{\tilde{\mathcal{M}}}(x)=y$ with good generalization performance on the test set of the target domain $\mathcal{X}^{test}_{target}$ by automatic counterfactual augmentation. For any sample $x$ with its label $y$, which consists of a token sequence $x=\{t_{cls}, t_0, t_1, ..., t_i, ..., t_{sep}\}$, a word-group $g$ is treated as a combination of any number of tokens in the sequence. Ideally, $g$ reflects the true causal feature of the sample. The goal of word-group mining is to provide a corresponding word-group set $\mathcal{G}_x$ for each sample. 

\subsection{Word-Group Mining}\label{sec:wg_mining}
Given the observations in Figure~\ref{fig:example}, we find that a single token does not cover the causal feature of the sample well in some cases, so we expect to use any combination of tokens, namely a word-group, to represent the real causal features. Theoretically, all the tokens in a sentence could be part of a word-group, but considering all the tokens would certainly complicate the search process. A wise pre-consideration is that the presence of some words in the sample, such as \textit{A} in Figure~\ref{fig:example}, will have a weak effect on the final prediction, so they can be easily eliminated to reduce the search space. This process is called candidate causal word mining. 

\subsubsection{Candidate Causal Words Mining}
We use a post-hoc interpretable method to analyze candidate causal words in each sample. It's based on a fine-tuned model $\mathcal{M}'$ on $\mathcal{X}^{train}_{source}$ and attributes the impact of each token on the model's prediction. Here, integrated gradient, a widely used post-hoc interpretable method is adopted to determine causal words in training samples~\cite{SundararajanTY17, SikdarBH20}. For a input sample $x$, the gradient of the $i^{th}$ token $t_i$ can be represented as:
\begin{equation}
	IG_{t_i} = (x_i - x_{\vec{0}})*\int_{0}^{1}\frac{\partial f_{\mathcal{M}'}(x_{\vec{0}}+\alpha*(x_i-x_{\vec{0}}))}{\partial x_i} d\alpha, 
\end{equation}
where $x_i$ denotes the embedding of $t_i$ with $d$ dimensions, $f_{\mathcal{M}'}(x)$ is the mapping function which maps $x$ to the corresponding label $y$ through the fine-tuned model $\mathcal{M}'$. $x_{\vec{0}}$ is a all-zero embedding. Subsequently, Riemann-sum approximation is used to approximate the gradient by summing small intervals along the straightline path from $x_i$ to $x_{\vec{0}}$:
\begin{equation}
	IG_{t_i} = (x_i - x_{\vec{0}})*\sum_{j=1}^{m}\frac{\partial f_{\mathcal{M}'}(x_{\vec{0}}+\frac{j}{m}*(x_i-x_{\vec{0}}))}{\partial x_i}\frac{1}{m}, 
\end{equation}
where $m$ is the number of steps in the Riemann-sum approximation which is set to 50 as adviced by Captum\footnote{https://github.com/pytorch/captum}. The L2 norm is then used to convert the gradient vector corresponding to each token into a scalar as the final attributing score $\Vert IG_{t_i} \Vert$. Since a token may appear multiple times in a sample, that is, $t_i$ may be the same as $t_j$, so we calculate the corpus-level attribution score corresponding to $w_{t_i}$ as:
\begin{equation}
	CS_{w_{t_i}} =\frac{1}{Freq(w_{t_i})}\sum_{j=1}^{Freq(w_{t_i})} \Vert IG_{w_{t_i}} \Vert_j,
\end{equation}
where $Freq(w_{t_i})$ is the total occurrences of $w_{t_i}$ in $\mathcal{X}^{train}_{source}$ and $w_{t_i} \in \mathcal{W}$ is the word of $t_i$ where $\mathcal{W}$ is the vocabulary of the training corpus. According to $CS_w$, a list of ranked causal words can be obtained, and we take the top 20\% tokens as the final candidate causal words $\tilde{\mathcal{W}}$. In this way, the number of tokens to be searched within a sample is reduced, reducing the complexity of the search.

\subsubsection{Word-Group Search}
Through the pre-selection of candidate causal words, each sample can obtain a causal word list $\tilde{\mathcal{W}}_x=\{w_0, w_1, ..., w_l\}, \tilde{\mathcal{W}}_x\subset \tilde{\mathcal{W}}$. Then, by searching for any combination of the tokens in $\tilde{\mathcal{W}}_x$ and estimating the causal effect of the combination, we hope to obtain a sorted set of word-groups $\mathcal{G}_x$. For this purpose, we propose an improved beam search Algorithm to search for word-groups with the greatest causal effects. Here, considering the counterfactual framework of causal inference~\cite{winship1999estimation}, the causal effect is defined as the disturbance effect to the probability distribution of a trained language model $\mathcal{M}'$ caused by automatic counterfactual augmentation against a word-group. 

For example, given the sample '\textit{A interesting and important film}' and one of its word-groups $\{interesting, important\}$, the corresponding automaticly counterfactual result is '\textit{A boring and unimportant film}', where the corresponding token is replaced by its antonym. If a token doesn't have an antonym, we adopt the lazy counterfactual appraoch~\cite{GargPLTCB19} and replace the token with LPLMs' mask token. The sample after the counterfactual augmentation is represented by $\bar{x}_g$. Correspondingly, the probability distributions of $\mathcal{M}'$ are $p(x)$ and $p(\bar{x}_g)$. To measure the agreement between the distributions, Jensen–Shannon Divergence (JSD)~\cite{guo2022auto}, a symmetric and smooth Kullback–Leibler divergence (KLD) is used:
\begin{equation}
	JSD_g = \frac{1}{2}KLD(p_(\bar{x}_g||p(x)) + \\ \frac{1}{2}KLD(p(x)||p(\bar{x}_g).
\end{equation}
The greater the value of $JSD_g$, the greater the impact of perturbations against word-group $g$, thus the more likely $g$ is to become a robust causal feature. 

\renewcommand{\algorithmicrequire}{\textbf{Input:}}
\renewcommand{\algorithmicensure}{\textbf{Output:}}
\begin{algorithm}[htb]  
	\caption{Word-Group Search Algorithm}  
	\label{alg:myalg}  
	\begin{algorithmic}[1]
		\REQUIRE Trained Language Model $\mathcal{M}'$, candidate causal words for current sample $\tilde{\mathcal{W}}_x$, max word-group length $L$ and beam width $K$. 
		\ENSURE  Generated sorted word-groups $\mathcal{G}_x$. 
		\STATE  $\mathcal{G}_x \leftarrow \{\}$ 
		\STATE  Current candidate word-groups $\mathcal{G}_{cand} \leftarrow \tilde{\mathcal{W}}_x$
		\FOR{$l\leftarrow 1$ to $L$}
			\STATE $\mathcal{G}_{gen} \leftarrow Sorted_{g\in \mathcal{G}_{cand}}(JSD(p(x)||p(\bar{x}_g))[:K]$  
			\STATE $\mathcal{G}_x \leftarrow \mathcal{G}_x \cap \mathcal{G}_{gen}$
			\STATE $\mathcal{G}_{cand} \leftarrow \{g\oplus w|\forall g\in \mathcal{G}_{gen}, \forall w \in \tilde{\mathcal{W}}_x, w\notin g \}$
		\ENDFOR
		\STATE $\mathcal{G}_x \leftarrow Sorted(\mathcal{G}_x)$
		\RETURN $\mathcal{G}_x[:L]$
	\end{algorithmic}  
\end{algorithm} 

Further, Algorithm~\ref{alg:myalg} summarizes the proposed word-groups search method. First, the algorithm retrieves the top $K$ tokens by causal effect from the candidate causal words $\tilde{\mathcal{W}}_x$ in Line $4$, where $[:K]$ represents the interception of the top $K$ items of the sorted array. Then, Algorithm~\ref{alg:myalg} takes these top $K$ tokens as basic word-groups with length 1, and continue to generate word-groups with length 2 in Line $6$. Here, $g\oplus w$ indicates extension of word-groups $g$ with new word $w$. In practice, we make sure that the new word $w$ does not exist in $g$. Then, generation continues on the basis of the new candidate word-groups $\mathcal{G}_{cand}$ in the next cycle in Line $4$, until new word-groups in the current circle reach the specified maximum length. Finally, we rank the causal effects of the generated word-groups (Line $8$) and select the top $L$ as true causal features (Line $9$). A simple example with $K=L=2$ can be found in Figure~\ref{fig:model}. In this paper, we adopt the configuration of $K=2$ and $L=3$, to reduce the complexity of search and ensure that reasonable word-groups are taken into account as much as possible.

\subsection{Multiple Causal Contrastive Learning}\label{sec:contrastive}
\textbf{Data augmentation based on word-groups}. After obtaining the word-groups $\mathcal{G}_x$, a special multiple contrastive learning framework is designed to make full use of the mining results~\cite{HjelmFLGBTB19}. For contrastive learning, an important premise is to obtain the corresponding positive and negative examples through data augmentation. For the negative samples, we get it via the automatic counterfactual substitution of word-groups. Since word-groups represent the most likely causal features, samples obtained by counterfactual are most likely to have opposite semantics. For positive samples, we can capture them by randomly perturbing the tokens that do not belong to word-groups. Specifically, we represent the composition of word-groups as $\mathcal{W}_{G_x}$, and mask $\tilde{\mathcal{W}}_x - \mathcal{W}_{G_x}$ randomly with a probability of 50\% as~\cite{MoonMLLS21}. Thus, the collection of the obtained augmented samples is written as $(x, x^+, x^-_{1}, ..., x^-_{L})$. 

\textbf{Multiple negative samples voting mechanism}. The negative samples correspond to the word-groups with different causal effect, so we expect the model to distinguish among them. Inspired by some research on collective decision making~\cite{el2019shared, geng2020collective}, the losses of multiple negative samples are combined to adaptively determine the contribution of each negative sample to the model optimization. Specifically, for the collection of augmented samples above, $\mathcal{M}'$ is easy to access to their corresponding representations as $(h, h^+, h^-_{1}, ..., h^-_{l})$. Mimicking SimCLR~\cite{ChenK0H20}, a simple MLP that shares parameters maps them to a lower dimensional representation space as $(z, z^+, z^-_{1}, ..., z^-_{l})$. 

Then, we design an attention-based adaptive voting module, which learns about the contributions of different word-groups as:
\begin{equation}
	\alpha_L = softmax(([z^-_{1}, ..., z^-_{l}])W + b),
\end{equation}
where $[, ..., ]$ represents the concatenation of the vectors, $W\in \mathcal{R}^{d_z*l}$ is the learnable weight parameter and $b\in \mathcal{R}^l$ denotes the bias where $d_z$ denotes the hidden dimension of $z$. $softmax$ is used to normalize the learned contributions. Subsequently, contrastive learning loss can be written as the following margin-based ranking loss~\cite{ZhangZLZH20}:
\begin{equation}
	\label{eq:cl}
	\mathcal{L}_{CL} = max(0, 
	\Delta + cos(z, z^+) - \alpha_l\odot [cos(z,z^-_1), ..., cos(z,z^-_l)]),
\end{equation}
where $\Delta$ is a margin value that we set to 1, $cos$ denotes the cosine similarity of the vectors, $\odot$ represents the Hadamard product of the vectors. Finally, the total loss function is the weighted sum of the cross entropy and the above loss:
\begin{equation}
	\label{eq:loss}
	\mathcal{L} = \mathcal{L}_{CE}+\lambda \mathcal{L}_{CL},
\end{equation}
where $\lambda$ is the weight that needs to be further explored and $\mathcal{L}_{CE}$ is the cross entropy loss on $\mathcal{X}_{source}^{train}$. 

\section{Datasets}\label{sec:datasets}
\begin{table}[htbp]
	\centering
	\setlength\tabcolsep{4pt}
	\renewcommand{\arraystretch}{0.8}
	\caption{The datasets and the corresponding partitioning. 0/1 denotes the number of negative and positive samples. }
	\begin{tabular}{lcccc}
		\hline
		Datasets & {books} & {dvd} & {electronics} & {kitchen} \\
		Train/Test &    2,000/4,465   &   2,000/3,586    &   2,000/5,681    &  2,000/5,954 \\
		0/1  & 3,201/3,264 & 2,779/2,807 & 3,824/3,857 & 3,991/3,954 \\ \hline
		Datasets & {mr} & {foods} & {sst2} & {kindle} \\
		Train/Test &    7,108/3,554   &   21,085/9,008   &   67,349/872    & 7,350/3,150 \\
		0/1  & 5,485/5,375 & 6,986/23,107  & 30,208/38,013 & 5,287/5,283  \\ \hline
		Datasets & {Davidson} & {OffEval} & {ToxicTweets} & {Abusive} \\
		Train/Test &    17,346/7,436   &   13,240/860    &   21,410/9,178    & 2,767/1,187 \\
		0/1  & 4,163/20,619 &   9,460/4,640   &    15,294/15,294   & 1,998/1,996  \\ \hline
	\end{tabular}%
	\label{tab:dataset}%
\end{table}%

To verify the validity of the proposed method, a variety of text classification tasks are explored on 12 different datasets. Specifically, the datasets can be divided into three groups as shown in Table~\ref{tab:dataset}.
\begin{itemize}
	\item Multi-Domain Sentiment Dataset\footnote{https://www.cs.jhu.edu/~mdredze/datasets}~\cite{BlitzerDP07}. It contains four different Amazon product reviews, \textbf{books}, \textbf{dvd}, \textbf{electronics} and \textbf{kitchen} which are contained in four different directories. We select \textit{positive.review} and \textit{negative.review} as the training set, and use \textit{unlabeled.review} as the test set. 
	\item More sources of sentiment classification datasets, including: Movie Review (\textbf{mr}) dataset containing binary categories~\cite{PangL05}. FineFood (\textbf{foods})~\cite{McAuleyL13} for food reviews scored on a scale from 1 to 5. Following~\cite{MoonMLLS21}, ratings $5$ are regarded as positive and ratings $1$ are regarded as negative. Stanford Sentiment Treebank (\textbf{sst2})~\cite{SocherPWCMNP13} with sentence binary classification task containing human annotations in movie reviews and their emotions. Kindle reviews (\textbf{kindle})~\cite{HeM16} from the Kindle Store, where each review is rated from 1 to 5. Following~\cite{WangC20, WangC21}, reviews with ratings $4, 5$ are positive and reviews with ratings $1, 2$ are negative.
	\item Toxic detection datasets including: \textbf{Davidson}~\cite{DavidsonWMW17} collected from Twitter which contains three categories, hate speech, offensive or not. \textbf{OffEval}~\cite{ZampieriMNRFK19} collected from Twitter which is divided into offensive and non-offensive. \textbf{ToxicTweets}\footnote{https://huggingface.co/datasets/mc7232/toxictweets} from Twetter, where toxic, severe toxic, obscene, threat, insult, and identity hate are marked. We chose toxic or not as our dichotomous task and obtain balanced categories by downsampling the non-toxic samples. \textbf{Abusive} from Kaggle\footnote{https://www.kaggle.com/datasets/hiungtrung/abusive-language-detection} for binary abusive language detection. We collectively treat offensive, hateful, abusive speech as toxic, and we convert toxic language detection as a binary text classification task. For all the toxic detection datasets, we delete non-English characters, web links, dates, and convert all the words to lowercase.
\end{itemize}

Then, we perform different tasks on a number of different baselines for the above datasets, including cross-domain text generalization, robustness testing against text attacks and gender fairness analysis. 

\definecolor{mygreen}{rgb}{ .886,  .937,  .855}
\begin{table*}[htbp]
	\centering
	\setlength\tabcolsep{3pt}
	\renewcommand{\arraystretch}{0.8}
	\caption{\textbf{BERT}'s results (accuracy \%) on cross-domain text classification. Bold indicates the optimal result, green indicates the average of the test results on different target domains with a fixed source domain. }
	\begin{tabular}{cccccccccccccc}
		\hline
		\multicolumn{2}{c}{Datasets} & \multicolumn{5}{c}{Models}            & \multicolumn{2}{c}{Datasets} & \multicolumn{5}{c}{Models} \\ \hline
		\multicolumn{1}{c}{Source} & Target & \multicolumn{1}{l}{BERT} & \multicolumn{1}{l}{AGC} & \multicolumn{1}{l}{MASKER} & \multicolumn{1}{l}{C2L} & \multicolumn{1}{l}{ACWG} & \multicolumn{1}{c}{Source} & Target & \multicolumn{1}{l}{BERT} & \multicolumn{1}{l}{AGC} & \multicolumn{1}{l}{MASKER} & \multicolumn{1}{l}{C2L} & \multicolumn{1}{l}{ACWG} \\ \hline
		\multirow{5}[0]{*}{books} & books & 81.27 & 80.25 & 81.85 & 82.39 & 82.68 & \multirow{5}[0]{*}{dvd} & books & 79.84 & 77.69 & 80.60 & 79.05 & 81.76 \\
		& dvd   & 80.51 & 79.55 & 80.00  & 79.66 & 81.04 &       & dvd   & 76.68 & 78.81 & 75.85 & 75.29 & 80.87 \\
		& electronics & 70.76 & 70.22 & 68.76 & 69.23 & 77.57 &       & electronics & 62.61 & 65.16 & 68.53 & 69.10  & 77.41 \\
		& kitchen & 78.27 & 78.13 & 79.94 & 78.12 & 79.31 &       & kitchen & 68.44 & 75.46 & 73.02 & 80.18 & 83.30 \\
		& \cellcolor{mygreen}Average & \cellcolor{mygreen}77.70 & \cellcolor{mygreen}77.04 & \cellcolor{mygreen}77.64 & \cellcolor{mygreen}77.35 & \cellcolor{mygreen}\textbf{80.15} & & \cellcolor{mygreen}Average & \cellcolor{mygreen}71.89 & \cellcolor{mygreen}74.28 & \cellcolor{mygreen}74.50 & \cellcolor{mygreen}75.91 & \cellcolor{mygreen}\textbf{80.84} \\ \hline
		\multirow{5}[0]{*}{electronics} & books & 83.92 & 79.47 & 77.68 & 83.92 & 85.74 & \multirow{5}[0]{*}{kitchen} & books & 84.61 & 81.33 & 87.16 & 85.52 & 87.96 \\
		& dvd   & 73.51 & 73.79 & 70.33 & 71.44 & 77.36 &       & dvd   & 67.32 & 73.92 & 68.96 & 73.21 & 76.57 \\
		& electronics & 75.40  & 80.04 & 76.85 & 80.23 & 78.30  &       & electronics & 69.74 & 77.16 & 73.47 & 76.73 & 78.50 \\
		& kitchen & 84.22 & 85.31 & 84.42 & 84.40  & 84.95 &       & kitchen & 81.20  & 83.35 & 81.26 & 83.80  & 83.86 \\
		& \cellcolor{mygreen}Average & \cellcolor{mygreen}79.26 & \cellcolor{mygreen}79.65 & \cellcolor{mygreen}77.32 & \cellcolor{mygreen}80.00 & \cellcolor{mygreen}\textbf{81.59} &  & \cellcolor{mygreen}Average & \cellcolor{mygreen}75.71 & \cellcolor{mygreen}78.94 & \cellcolor{mygreen}77.71 & \cellcolor{mygreen}79.82 & \cellcolor{mygreen}\textbf{81.72} \\ \hline
		\multirow{5}[0]{*}{mr} & mr    & 85.51 & 85.14 & 84.81 & 85.37 & 85.34 & \multirow{5}[0]{*}{foods} & mr    & 61.31 & 65.44 & 60.79 & 65.14 & 68.35 \\
		& foods & 69.95 & 75.61 & 60.95 & 77.48 & 83.95 &       & foods & 96.40  & 96.46 & 96.12 & 96.20  & 96.29 \\
		& sst2  & 91.97 & 92.32 & 92.43 & 91.40  & 91.74 &       & sst2  & 70.64 & 72.71 & 72.33 & 73.97 & 76.38 \\
		& kindle & 84.06 & 84.60  & 85.14 & 84.79 & 85.97 &       & kindle & 72.44 & 76.57 & 77.46 & 77.05 & 78.76 \\
		& \cellcolor{mygreen}Average & \cellcolor{mygreen}82.87 & \cellcolor{mygreen}84.42 & \cellcolor{mygreen}80.83 & \cellcolor{mygreen}84.76 & \cellcolor{mygreen}\textbf{86.75} &       & \cellcolor{mygreen}Average & \cellcolor{mygreen}75.20 & \cellcolor{mygreen}77.80 & \cellcolor{mygreen}76.68 & \cellcolor{mygreen}78.09 & \cellcolor{mygreen}\textbf{79.95} \\ \hline
		\multirow{5}[0]{*}{sst2} & mr    & 87.76 & 88.04 & 87.59 & 88.12 & 88.76 & \multirow{5}[0]{*}{kinde} & mr    & 80.47 & 81.40  & 79.40  & 80.39 & 81.04 \\
		& foods & 81.01 & 82.24 & 82.65 & 81.56 & 86.45 &       & foods & 83.47 & 83.98 & 82.76 & 86.00    & 84.62 \\
		& sst2  & 91.74 & 91.85 & 91.74 & 92.20  & 91.86 &       & sst2  & 86.47 & 84.06 & 84.14 & 85.09 & 86.94 \\
		& kindle & 85.11 & 85.36 & 85.87 & 85.81 & 85.21 &       & kindle & 89.21 & 89.29 & 89.43 & 89.11 & 89.52 \\
		& \cellcolor{mygreen}Average & \cellcolor{mygreen}86.41 & \cellcolor{mygreen}86.87 & \cellcolor{mygreen}86.96 & \cellcolor{mygreen}86.92 & \cellcolor{mygreen}\textbf{88.07} &       & \cellcolor{mygreen}Average & \cellcolor{mygreen}84.91 & \cellcolor{mygreen}84.68 & \cellcolor{mygreen}83.93 & \cellcolor{mygreen}85.15 & \cellcolor{mygreen}\textbf{85.53} \\ \hline
		\multirow{5}[0]{*}{Davidson} & Davidson & 96.46 & 96.48 & 96.06 & 96.41 & 96.32 & \multirow{5}[0]{*}{OffEval} & Davidson & 82.23 & 82.41 & 82.52 & 83.58 & 83.84 \\
		& OffEval & 79.53 & 80.70  & 80.35 & 80.47 & 80.81 &       & OffEval & 83.72 & 85.35 & 82.33 & 83.07 & 84.53 \\
		& Abusive & 76.91 & 77.78 & 77.76 & 80.20  & 79.11 &       & Abusive & 80.79 & 83.15 & 80.88 & 85.35 & 83.07 \\
		& ToxicTweets & 74.72 & 79.62 & 81.09 & 80.21 & 82.61 &       & ToxicTweets & 82.79 & 87.49 & 85.25 & 88.14 & 88.98 \\
		& \cellcolor{mygreen}Average & \cellcolor{mygreen}81.91 & \cellcolor{mygreen}83.65 & \cellcolor{mygreen}83.82 & \cellcolor{mygreen}84.32 & \cellcolor{mygreen}\textbf{84.70} &       & \cellcolor{mygreen}Average & \cellcolor{mygreen}82.38 & \cellcolor{mygreen}84.60 & \cellcolor{mygreen}82.75 & \cellcolor{mygreen}85.04 & \cellcolor{mygreen}\textbf{85.11} \\ \hline
		\multirow{5}[0]{*}{Abusive} & Davidson & 81.86 & 84.44 & 83.31 & 82.37 & 84.37 & \multirow{5}[0]{*}{ToxicTweets} & Davidson & 87.80  & 85.92 & 86.97 & 86.66 & 86.51 \\
		& OffEval & 78.72 & 78.14 & 77.79 & 77.84 & 80.91 &       & OffEval & 79.42 & 81.83 & 77.79 & 82.14 & 82.56 \\
		& Abusive & 92.41 & 94.36 & 93.68 & 93.58 & 94.69 &       & Abusive & 79.11 & 82.20  & 77.17 & 81.13 & 83.15 \\
		& ToxicTweets & 85.47 & 85.27 & 85.21 & 86.31 & 85.86 &       & ToxicTweets & 91.89 & 92.51 & 91.27 & 92.43 & 92.62 \\
		& \cellcolor{mygreen}Average & \cellcolor{mygreen}84.62 & \cellcolor{mygreen}85.55 & \cellcolor{mygreen}85.00 & \cellcolor{mygreen}85.03 & \cellcolor{mygreen}\textbf{86.46} &       & \cellcolor{mygreen}Avurage & \cellcolor{mygreen}84.56 & \cellcolor{mygreen}85.62 & \cellcolor{mygreen}83.30 & \cellcolor{mygreen}85.59 & \cellcolor{mygreen}\textbf{86.21} \\ \hline
	\end{tabular}%
	\label{tab:bert_results}%
\end{table*}%

\section{Tasks and Experimental Results}
In this section, we introduce experimental results on the corresponding datasets to address the following key questions:
\begin{itemize}
	\item \textbf{Q1}. Does ACWG help to generalize LPLMs?
	\item \textbf{Q2}. Does ACWG improve the robustness of LPLMs?
	\item \textbf{Q3}. Does ACWG improve the fairness of LPLMs?
	\item \textbf{Q4}. How does the proposed Word-Groups-based mining approach help improve the performance of different tasks?
\end{itemize}

\subsection{Q1: In-domain and Cross-domain Text Classificaion}\label{sec:cross-domain}
To answer \textbf{Q1}, we test the OOD generalization performance of different datasets and further explore the experimental parameters. 

\subsubsection{Baselines and Details}

\textbf{Baselines}. The cross-domain generalization is verified by training on the source domain and testing on the target domain. They have different data distributions. Several different shortcut mitigation or automatic counterfactual agumentation approaches are compared. \textbf{Automatically Generated Counterfactuals (AGC)}~\cite{WangC21}, which augments the training data with automatically generated counterfactual data by substituting causal features with the antonyms and assigning the opposite labels. Then, the augmented samples are added to the training dataset to train a robust model. \textbf{MASKER}~\cite{MoonMLLS21}, which improves the cross-domain generalization of language models through the keyword shortcuts reconstruction and entropy regularization. It uses tokens with high LPLMs attention scores as possible shortcuts. \textbf{C2L}~\cite{ChoiJHH22}, which monitors the causality of each word collectively through a set of automatically generated counterfactual samples and uses contrastive learning to improve the robustness of the model.

\textbf{Details}. As our main experiment, we conduct training on the training set of the source domain $\mathcal{X}_{source}^{train}$, and save the optimal models which have the best results on $\mathcal{X}_{source}^{test}$. Then, the optimal models are used to perform text attack testing and fairness testing. The batch of all datasets and all baselines is uniformly set to 64, and the learning rate is $1e-5$. We set epoch to 5 and use Adam as the optimizer. All the codes are written using pytorch and trained on four NVIDIA A40 GPUs. For the baselines, officially published codes are used to replicate the experimental results. For AGC, we identify the causal features by picking the closest opposite matches which have scores greater than 0.95 as suggested in the original paper. For MASKER, we set the weights of the two regularization terms to 0.001 and 0.0001 for cross-domain generalization. For C2L, we set the number of positive/negative pairs for comparison learning to 1, and search for the optimal weight of contrastive learning loss in $[0.1, 0.7, 1.0]$. All the key parameters of the baselines are consistent with those reported in the original paper. 

\definecolor{mygreen}{rgb}{ .886,  .937,  .855}
\begin{table*}[htbp]
	\centering
	\setlength\tabcolsep{3pt}
	\renewcommand{\arraystretch}{0.8}
	\caption{\textbf{RoBERTa}'s results (accuracy \%) on cross-domain text classification. Bold indicates the optimal result, green indicates the average of the test results on different target domains with a fixed source domain. }
	\begin{tabular}{cccccccccccccc}
		\hline
		    \multicolumn{2}{c}{Datasets} & \multicolumn{5}{c}{Models}            & \multicolumn{2}{c}{Datasets} & \multicolumn{5}{c}{Models} \\
		\multicolumn{1}{c}{Source} & Target & \multicolumn{1}{l}{RoBERTa} & \multicolumn{1}{l}{AGC} & \multicolumn{1}{l}{MASKER} & \multicolumn{1}{l}{C2L} & \multicolumn{1}{l}{ACWG} & \multicolumn{1}{c}{Source} & Target & \multicolumn{1}{l}{BERT} & \multicolumn{1}{l}{AGC} & \multicolumn{1}{l}{MASKER} & \multicolumn{1}{l}{C2L} & \multicolumn{1}{l}{ACWG} \\ \hline
		\multirow{5}[0]{*}{books} & books & 84.37 & 84.30  & 84.49 & 82.82 & 84.57 & \multirow{5}[0]{*}{dvd} & books & 73.01 & 83.09 & 77.98 & 82.52 & 82.75 \\
		& dvd   & 80.37 & 83.35 & 79.77 & 82.13 & 82.82 &       & dvd   & 85.40  & 84.66 & 84.51 & 84.47 & 85.30 \\
		& electronics & 68.88 & 75.06 & 70.21 & 79.33 & 78.49 &       & electronics & 67.61 & 70.03 & 66.33 & 72.24 & 79.44 \\
		& kitchen & 82.44 & 84.19 & 82.46 & 83.29 & 84.26 &       & kitchen & 73.52 & 81.09 & 81.05 & 82.39 & 82.15 \\
		& \cellcolor{mygreen}Average & \cellcolor{mygreen}79.02 & \cellcolor{mygreen}81.73 & \cellcolor{mygreen}79.23 & \cellcolor{mygreen}81.89 & \cellcolor{mygreen}\textbf{82.54} &       & \cellcolor{mygreen}Average & \cellcolor{mygreen}74.89 & \cellcolor{mygreen}79.72 & \cellcolor{mygreen}77.47 & \cellcolor{mygreen}80.41 & \cellcolor{mygreen}\textbf{82.41} \\ \hline 
		\multirow{5}[0]{*}{electronics} & books & 69.54 & 79.99 & 72.03 & 78.45 & 79.35 & \multirow{5}[0]{*}{kitchen} & books & 69.41 & 77.11 & 69.74 & 77.87 & 80.49 \\
		& dvd   & 70.36 & 80.63 & 70.58 & 80.90  & 81.76 &       & dvd   & 75.38 & 78.57 & 78.05 & 78.75 & 82.37 \\
		& electronics & 83.70  & 85.64 & 85.01 & 85.63 & 86.36 &       & electronics & 75.48 & 82.99 & 77.89 & 82.78 & 83.98 \\
		& kitchen & 76.05 & 85.16 & 76.32 & 85.35 & 86.48 &       & kitchen & 87.68 & 87.24 & 88.13 & 87.59 & 88.75 \\
		& \cellcolor{mygreen}Average & \cellcolor{mygreen}74.91 & \cellcolor{mygreen}82.86 & \cellcolor{mygreen}75.99 & \cellcolor{mygreen}82.58 & \cellcolor{mygreen}\textbf{83.49} &       & \cellcolor{mygreen}Average & \cellcolor{mygreen}76.99 & \cellcolor{mygreen}81.48 & \cellcolor{mygreen}78.45 & \cellcolor{mygreen}81.75 & \cellcolor{mygreen}\textbf{83.90} \\ \hline
		\multirow{5}[0]{*}{mr} & mr    & 87.96 & 88.89 & 88.60  & 89.39 & 89.05 & \multirow{5}[0]{*}{foods} & mr    & 65.53 & 72.40  & 71.53 & 77.57 & 80.39 \\
		& foods & 78.13 & 86.29 & 75.26 & 85.44 & 85.90  &       & foods & 94.25 & 97.20  & 93.93 & 97.32 & 97.21 \\
		& sst2  & 93.23 & 93.12 & 93.35 & 92.89 & 93.69 &       & sst2  & 75.00    & 74.66 & 77.88 & 81.77 & 84.52 \\
		& kindle & 87.90  & 88.00    & 89.05 & 88.95 & 89.02 &       & kindle & 77.78 & 77.30  & 79.68 & 81.81 & 84.29 \\
		& \cellcolor{mygreen}Average & \cellcolor{mygreen}86.81 & \cellcolor{mygreen}89.08 & \cellcolor{mygreen}86.57 & \cellcolor{mygreen}89.17 & \cellcolor{mygreen}\textbf{89.42} &       & \cellcolor{mygreen}Average & \cellcolor{mygreen}78.14 & \cellcolor{mygreen}80.39 & \cellcolor{mygreen}80.76 & \cellcolor{mygreen}84.62 & \cellcolor{mygreen}\textbf{86.60} \\ \hline
		\multirow{5}[0]{*}{sst2} & mr    & 89.59 & 89.11 & 89.05 & 89.76 & 89.95 & \multirow{5}[0]{*}{kinde} & mr    & 83.03 & 82.89 & 82.36 & 82.40  & 85.31 \\
		& foods & 79.73 & 90.23 & 80.00    & 86.78 & 90.16 &       & foods & 79.98 & 84.72 & 82.51 & 83.02 & 89.08 \\
		& sst2  & 94.15 & 94.04 & 93.69 & 94.15 & 95.30  &       & sst2  & 87.50  & 87.65 & 87.27 & 87.67 & 88.53 \\
		& kindle & 88.00 & 87.02 & 87.56 & 87.40  & 88.62 &       & kindle & 90.38 & 90.89 & 91.02 & 91.86 & 90.67 \\
		& \cellcolor{mygreen}Average & \cellcolor{mygreen}87.87 & \cellcolor{mygreen}90.10 & \cellcolor{mygreen}87.56 & \cellcolor{mygreen}89.52 & \cellcolor{mygreen}\textbf{91.01} &       & \cellcolor{mygreen}Average & \cellcolor{mygreen}85.22 & \cellcolor{mygreen}86.54 & \cellcolor{mygreen}85.79 & \cellcolor{mygreen}86.24 & \cellcolor{mygreen}\textbf{88.40} \\ \hline
		\multirow{5}[0]{*}{Davidson} & Davidson & 95.75 & 96.26 & 96.30  & 96.32 & 96.02 & \multirow{5}[0]{*}{OffEval} & Davidson & 85.80  & 83.93 & 84.75 & 82.19 & 84.44 \\
		& OffEval & 79.88 & 80.12 & 78.84 & 79.65 & 80.47 &       & OffEval & 81.98 & 83.14 & 83.26 & 84.77 & 84.88 \\
		& Abusive & 78.94 & 80.50  & 79.78 & 80.62 & 81.72 &       & Abusive & 79.44 & 81.80  & 80.54 & 84.76 & 84.84 \\
		& ToxicTweets & 82.37 & 83.60  & 80.74 & 82.40  & 84.51 &       & ToxicTweets & 88.60  & 87.79 & 88.15 & 88.03 & 88.8 \\
		& \cellcolor{mygreen}Average & \cellcolor{mygreen}84.24 & \cellcolor{mygreen}85.12 & \cellcolor{mygreen}83.92 & \cellcolor{mygreen}84.75 & \cellcolor{mygreen}\textbf{85.68} &       & \cellcolor{mygreen}Average & \cellcolor{mygreen}83.96 & \cellcolor{mygreen}84.17 & \cellcolor{mygreen}84.18 & \cellcolor{mygreen}84.94 & \cellcolor{mygreen}\textbf{85.74} \\ \hline
		\multirow{5}[0]{*}{Abusive} & Davidson & 82.44 & 83.23 & 81.68 & 82.15 & 83.39 & \multirow{5}[0]{*}{ToxicTweets} & Davidson & 87.21 & 87.13 & 86.75 & 87.06 & 87.52 \\
		& OffEval & 76.51 & 80.12 & 81.51 & 79.93 & 80.58 &       & OffEval & 78.95 & 80.00    & 78.60  & 80.70  & 82.09 \\
		& Abusive & 93.60  & 92.78 & 92.75 & 91.56 & 95.11 &       & Abusive & 78.52 & 78.85 & 81.30  & 78.43 & 81.37 \\
		& ToxicTweets & 81.76 & 84.31 & 82.53 & 88.22 & 89.69 &       & ToxicTweets & 93.40  & 94.09 & 93.15 & 94.49 & 94.59 \\
		& \cellcolor{mygreen}Average & \cellcolor{mygreen}83.58 & \cellcolor{mygreen}85.11 & \cellcolor{mygreen}84.62 & \cellcolor{mygreen}85.47 & \cellcolor{mygreen}\textbf{87.19} &       & \cellcolor{mygreen}Avurage & \cellcolor{mygreen}84.52 & \cellcolor{mygreen}85.02 & \cellcolor{mygreen}84.95 & \cellcolor{mygreen}85.17 & \cellcolor{mygreen}\textbf{86.39} \\
		\hline
	\end{tabular}%
	\label{tab:roberta_results}%
\end{table*}%

\subsubsection{Comparisons With State-of-the-Arts}
The experimental results of BERT and RoBERTa, two commonly used LPLMs, are reported in Table~\ref{tab:bert_results} and Table~\ref{tab:roberta_results}. In general, ACWG is able to achieve the best average in all cases, and has similar performance with BERT and RoBERTa as the backbones. First, we note that the attention-based shortcuts extraction method MASKER is not always effective. For example, compared to basic BERT, MAKSER shows degradation of performance on electronics, mr, ToxicTweets, and etc. This shows that attention score may not be suitable for robust feature extraction, and also indicates the importance of reasonable keyword mining methods. In contrast, counterfactual augmentation based methods AGC and C2L both achieve better results in most cases. But the former is superior to C2L in only a few cases, because it also includes samples with opposite augmentation as part of the training dataset, which is easily affected by the quality of the augmentation samples. While C2L adopts the form of contrastive learning and uses collaborative decision making to give a more robust counterfactual augmentative utilization. Finally, the proposed ACWG can obtain optimal values on all datasets, which indicates that mining word-groups and reasonably using them to generate counterfactual augmentation can stimulate LPLMs' ability to learn robust features, and therefore contribute to LPLMs' generalization. 

Due to the similarity of BERT's and RoBERTa's results, we take took BERT as the backbone to conduct in-depth exploration in the follow-up experiments.

\subsubsection{Parameters Exploration}
Two main parameters explored in relation to ACWG are the loss weight of comparative learning $\lambda$ and the number of word-groups used $l$.

\begin{figure*}[t!]
	\begin{center}
		\includegraphics[width=0.95\textwidth]{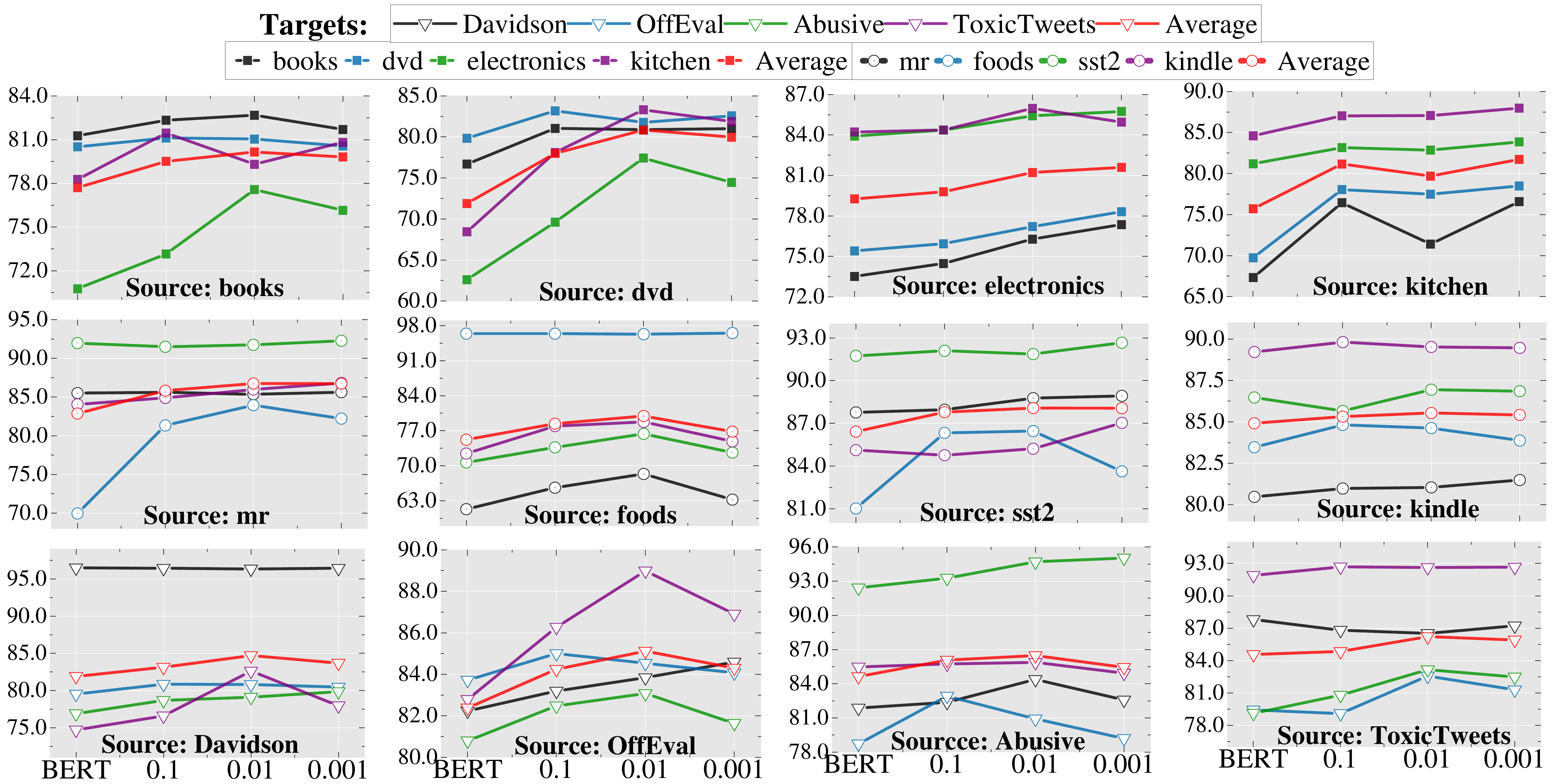}
		\label{fig:lambda}
		\caption{The optimum $\lambda$ for different datasets and the comparison with BERT. The dataset titles under different subgraphs represent the source domains for training, the datasets in the legends represent the target domains for test. Different combinations of colors and symbols represent different datasets. }
	\end{center}
\end{figure*} 

\textbf{Contrastive learning loss $\lambda$}. First, $\lambda$ in Eq.~\ref{eq:loss} is analyzed to determine the loss ratio of assisted contrastive learning. Since the optimal parameters of different datasets are difficult to be selected uniformly, our goal is to investigate the optimal magnitude of $\lambda$. Specifically, we show cross-domain generalization results in all cases and average performance changes for $\lambda\in \{0.1, 0.01, 0.001\}$ in Figure~\ref{fig:lambda} compared with BERT. Although there are differences among different datasets, the best results are produced at $0.01$ or $0.001$ for all averages. Therefore, we choose 0.01 or 0.001 as the optimal $\lambda$ value. In addition, in most cases, no matter the value of $\lambda$, ACWG is better than backbone, which further verifies the effectiveness of the proposed method. 

\begin{figure}[t!]
	\begin{center}
		\includegraphics[width=0.5\textwidth]{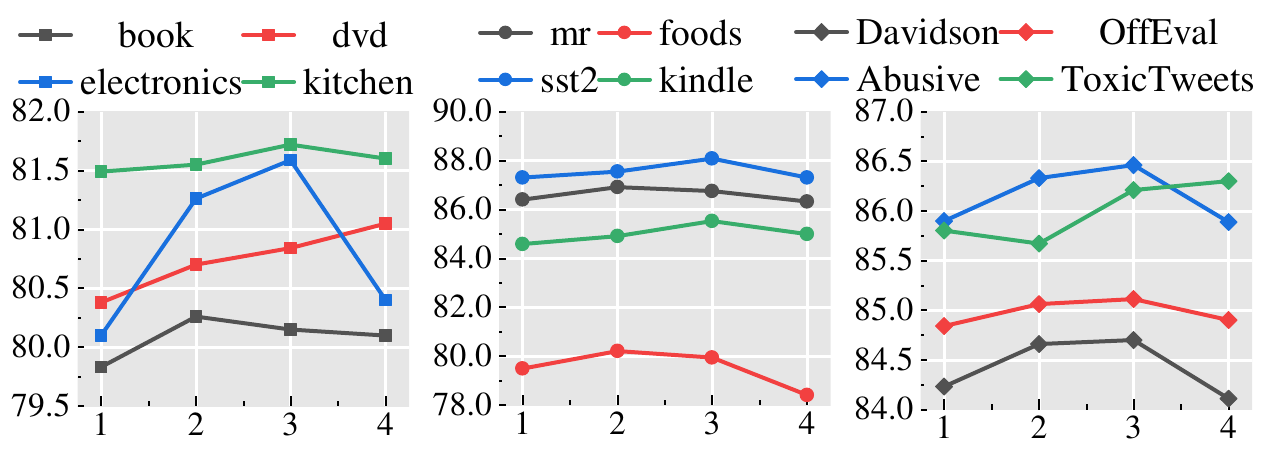}
		\label{fig:L}
		\caption{The average value of the target domain generalization effect on different dataset groups varies with the number of word-groups $l$. }
	\end{center}
\end{figure} 
\textbf{Word-groups number $l$}. Subsequently, $l\in \{1,2,3,4\}$ is further analyzed to determine a reasonable number of word-groups in Figure~\ref{fig:L}. Here, the average results of the target domain under each particular source domain are reported, because the results vary widely across different target domains, a comprehensive evaluation is used as the main decision basis, same as Figure~\ref{fig:lambda}. We note the inadequacy of a single word-group as it has a low tolerance for noise compared to the collective decision-making of multiple word-groups. But this does not mean that more word-groups will bring better results, because with the increase of word-groups, groups with lower causal effect will be included in the decision-making group, which will also introduce potential noise. As a result, the optimal results of most datasets are generated at 2 or 3, except for dvds and ToxicTweets. Therefore, we choose a stable value $l=3$ as the parameters for all datasets, even though this parameter may not represent the optimal results.

\subsubsection{Ablation Study} \label{sec:abl}
\begin{figure}[t!]
	\begin{center}
		\includegraphics[width=0.45\textwidth]{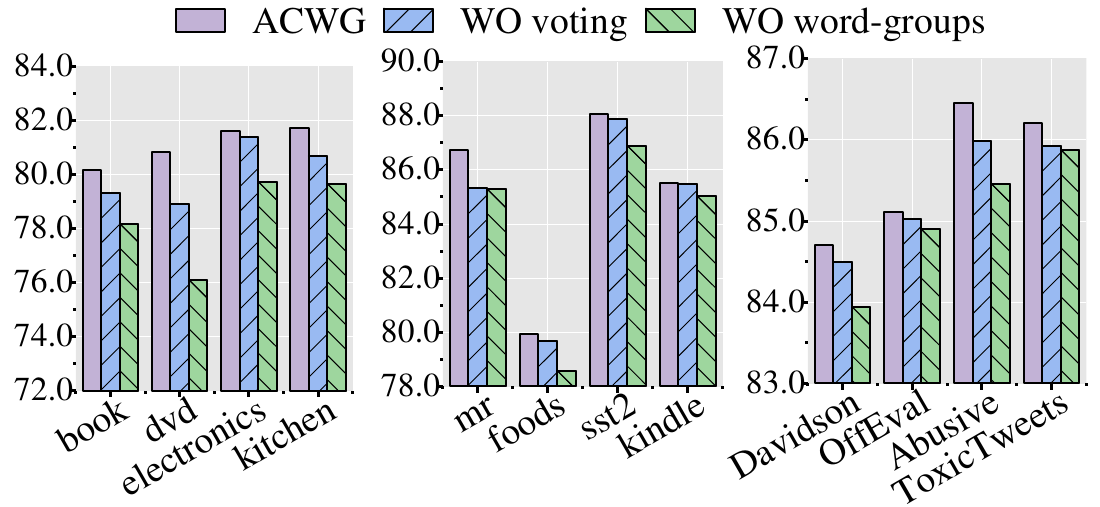}
		\label{fig:ab}
		\caption{The average performance under different source domains for ablation study. 'WO' denotes 'without'.}
	\end{center}
\end{figure} 
Ablation experiments are performed to analyze the effectiveness of the proposed key components in Figure~\ref{fig:ab} to further answer \textbf{Q1}. Specifically, two ACWG variants are considered: \textbf{WO voting} and \textbf{WO word-groups}. The former deletes the word-groups voting mechanism in Section~\ref{sec:contrastive}, that is, the word-group with the highest score is used to calculate directly. The latter means that the word-group search method proposed is not used, and only the keywords with the greatest causal effect are used for automatic counterfactual substitution. We observe that for all datasets, both variants of ACWG result in performance degradation. Among them, \textbf{WO voting} cause a smaller decline than \textbf{WO word-groups}, which indicates that word-groups mining is the main cause of ACWG performance improvement. The voting mechanism is based on word-groups, so the performance degradation of \textbf{WO voting} is lower. 

\subsection{Q2: Text Attack}\label{sec:attack}
To further verify the robustness of the proposed method (to answer \textbf{Q2}), several basic text attack methods are used to destroy the original text.

\textbf{Approaches}. \textbf{Probability Weighted Word Saliency (PWWS)}~\cite{RenDHC19}, a greedy algorithm including a new word substitution strategy by both the word saliency and the classification probability. \textbf{TextBugger}~\cite{LiJDLW19} finds the most important sentence, and uses a scoring function to look for keywords in the sentence, and then attacks the keywords. \textbf{TextFooler}~\cite{JinJZS20} looks for the key words that contribute the most to the sentence prediction by deleting words in sequence, and attacks the text by replacing the key words. 

\textbf{Details}. We attack the test sets of all the above datasets except for Multi-Domain Sentiment Dataset, then tests the performance of different models on the test sets after the attack. However, attacks on tokens often act on more than one tokens in the sample, so to prevent the semantics of the sample from changing too much due to the attacks, a constraint is added to limit the number of tokens to be attacked to $K$. But for Multi-Domain Sentiment Dataset, due to their long text length, the search time required for word replacement is estimated to be more than 24 hours on the 4*NVADIA A40 GPUs, so they are not discussed further.

\begin{figure*}[t!]
	\begin{center}
		\includegraphics[width=0.9\textwidth]{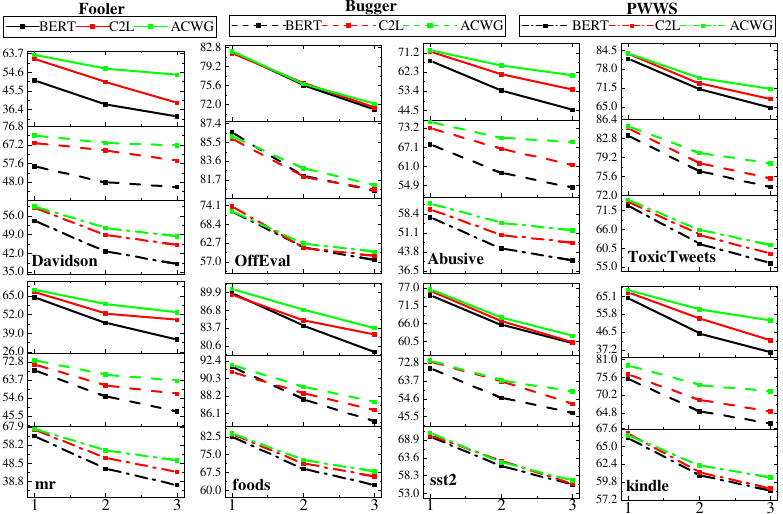}
		\label{fig:attack_plot}
		\caption{When the number of attack words is different (1,2,3), the three attack methods lead to the performance changes of BERT, C2L and PWWS. Different groups of graphs show the results of the same data set, with different colors representing different fine-tuned models. Different styles of polylines represent different attack methods, different subgraphs represent different datasets. }
	\end{center}
\end{figure*}

\textbf{Attack Results}. We report on the response of different models to attacks on the test sets in Figure~\ref{fig:attack_plot}. The effectiveness of the different attack methods is demonstrated because they perturb the sample by retrieving the most important words. As a result, we observe significant performance degradations due to text attacks on 8 datasets, especially as the number of words being attacked increases. But in most cases, a robust model can increase resistance to attacks, whether word-based C2L or word-groups-based ACWG. Furthermore, word-groups-based approach is more effective at resisting attacks than single-word-based model because word-groups contain a more rational causal structure and are more diverse. ACWG shows a trend where the advantage over BERT increases as the number of words attacked increases. This is also due to ACWG's learning of word-groups, which makes it more robust when dealing with multiple attacked words. 

\subsection{Q3: Gender Fairness}\label{sec:fairness}
Furthermore, although our method does not specifically study fairness on minority groups, such as gender and race, robust feature learning still helps to alleviate the bias of the model~\cite{YaoCYJR21}. To verify this idea and answer \textbf{Q3}, in this paper, we explore the gender bias that has been extensively studied by a set of gender attribute terms given by~\cite{NadeemBR20}. If a sample contains any of the keywords in the gender attribute terms, then we assume that the sample is likely to have gender unfairness. We screen potential gender bias samples in the test sets of Davidson and ToxicTweets since they have more samples.

\textbf{Fairness Metrics}. Furthermore, although our method does not specifically study fairness on minority groups, such as gender and race, robust feature learning still helps to alleviate the bias of the model~\cite{YaoCYJR21}. To verify this idea and answer \textbf{Q3}, in this paper, we explore the gender bias that has been extensively studied by a set of gender attribute terms given by~\cite{NadeemBR20}. If a sample contains any of the keywords in the gender attribute terms, then we assume that the sample is likely to have gender unfairness. We screen potential gender bias samples in the test sets of Davidson and ToxicTweets since they have more samples. Subsequently, the trained model is used to test fairness on the above subsets, using the following metrics. \textbf{Perturbation Consistency Rate (PCR)}. PCR is used to assess the robustness of the model to the gender perturbation of the sample, which measures the percentage of predicted results that have not changed if a gender attribute term in a sample is replaced with the opposite word. For example, if a sample '\textit{She is a good girl}' is predicted by the model as positive, then its gender perturbation sample '\textit{He is a good boy}' should have the same prediction result. If the results are different, the model may be gender-sensitive and make unfair judgments about \textit{She} and \textit{He}. \textbf{False Positive Equality Difference (FPED) and False Negative Equality Difference (FNED)}~\cite{ZhangBZBZZ20}. They are relaxations of Equalized Odds (also known as Error Rate Balance) defined by~\cite{BorkanDSTV19} as follows:
	\begin{equation}
		\begin{split}
			FPED = \sum_{z}^{}|FPR_z - FPR_{all}|, \\
			FNED = \sum_{z}^{}|FNR_z - FNR_{all}|,
		\end{split}
	\end{equation}
	where $FPR_{all}$ and $FNR_{all}$ denotes False Positive Rate and False Negative Rate in the whole test set, $FPR_z$ and $FNR_z$ represents the results in the corresponding gender group $z$, where $z=\{male, female\}$. The lower their values, the more fair the model is. 

\begin{table}[htbp]
	\centering
	\caption{Measurement results of gender fairness compared with BERT on Davidson and ToxicTweets. $\uparrow$ indicates that the smaller the value, the higher the fairness, while $\downarrow$ is opposite. }
	\begin{tabular}{lrrrr}
		\hline \hline
		\renewcommand{\arraystretch}{0.8}
		& \multicolumn{2}{c}{Davidson} & \multicolumn{2}{c}{ToxicTweets} \\ 
		& {BERT} & {ACWG} & {BERT} & {ACWG} \\ \hline
		PCR ($\% \uparrow$)   & 99.10  & 99.51 & 99.04 & 99.22 \\
		FPED ($\downarrow$)  & 0.0201 & 0.0116 & 0.0228 & 0.0181 \\
		FNED ($\downarrow$)  & 0.1441 & 0.0949 & 0.0406 & 0.0389 \\ \hline \hline
	\end{tabular}%
	\label{tab:bias}%
\end{table}%
\textbf{Fairness Results}. The measurement of fairness is reported in Table~\ref{tab:bias}. For PCR, ACWG outperforms BERT on both Davidson and ToxicTweets, indicating that the proposed method is more stable when flipping the attributes, without misjudgment due to differences between male and female. In addition, the lower FPED and FNED also indicate that ACWG made more balanced predictions for the male/female samples, further verifying its fairness. ACWG's fairness also stems from a more explicit causal feature reflected in word-groups, since gender is not the actual cause of the model's predictions, and it is easy for ACWG to exclude the influence of such non-causal features.

\newcommand{\mybox}[1]{\fcolorbox{black}{yellow!30}{#1}}
\begin{table*}[htbp]
	\renewcommand{\arraystretch}{0.5}
	\centering
	\caption{Case studies from different datasets. The yellow text box shows the word-group with the highest causal effect score.}
	\begin{tabular}{cp{32.6em}p{15em}c} \hline
		Datasets & \multicolumn{1}{c}{Text} & \multicolumn{1}{c}{Word-groups l=3} & Category \\  \hline
		sst2  & that \mybox{loves} its characters and communicates something rather \mybox{beautiful} about human nature & beautiful loves; and beautiful loves; beautiful loves something & positive \\ \hline
		foods & this coffee is strong but \mybox{no} flavor, \mybox{no} taste, \mybox{no} aroma. poor choice do not try. i would not \mybox{recommend} to \mybox{purchase}. & no purchase recommend; choice no recommend; no recommend & negative \\ \hline
		Abusive & ben affleck is the best those other people who are aguring with them scream that old guy with the gray hair looks like donald trumps \mybox{asshole} and \mybox{screw} the other guy beat him looks like gru with hair that long \mybox{ass} ugly nose ben affleck is the best defending muslims even when he is not its just the best brings a tear to my eyeawata & ass asshole screw; asshole beat screw; asshole screw & toxic \\ \hline
		Davidson & i got some lightskin \mybox{pussy} one time and the \mybox{bitch} \mybox{damn} near had me bout to propose. had some i had to immediately & bitch damn pussy; bitch damn i; bitch pussy & toxic \\ \hline
	\end{tabular}%
	\label{tab:case}%
\end{table*}%

\subsection{Q4: Label Flipping Rate}\label{sec:fr}
Similiar to~\cite{WenZZ0H22}, we want to measure the quality of the sample generated by the automatic counterfactual. If a counterfactual sample produces an effect, then it should result in an oppositely labeled sample, compared to the ground truth label. Further, this opposite sample can be used to enrich the train data and induce the model to consider word-groups that represent robust features. Therefore, \textbf{Label Flipping Rate} (LFR) is adopted to measure the effectiveness of generating counterfactual data. It is defined as the ratio at which the counterfactual flipping of the sample will predict a different result compared to the ground truth label:
\begin{equation}
	LFR = 1-\frac{\sum_{x\in \mathcal{X}} \Xi (y=argmax(p(\bar{x})))}{|\mathcal{X}|},
\end{equation}
where $\mathcal{X}$ is the data set on which the counterfactual is executed, $y$ represents the ground truth label of the sample corresponding to $x$, and $\Xi$ is the indicator function. 
\begin{figure}[t!]
	\begin{center}
		\includegraphics[width=0.45\textwidth]{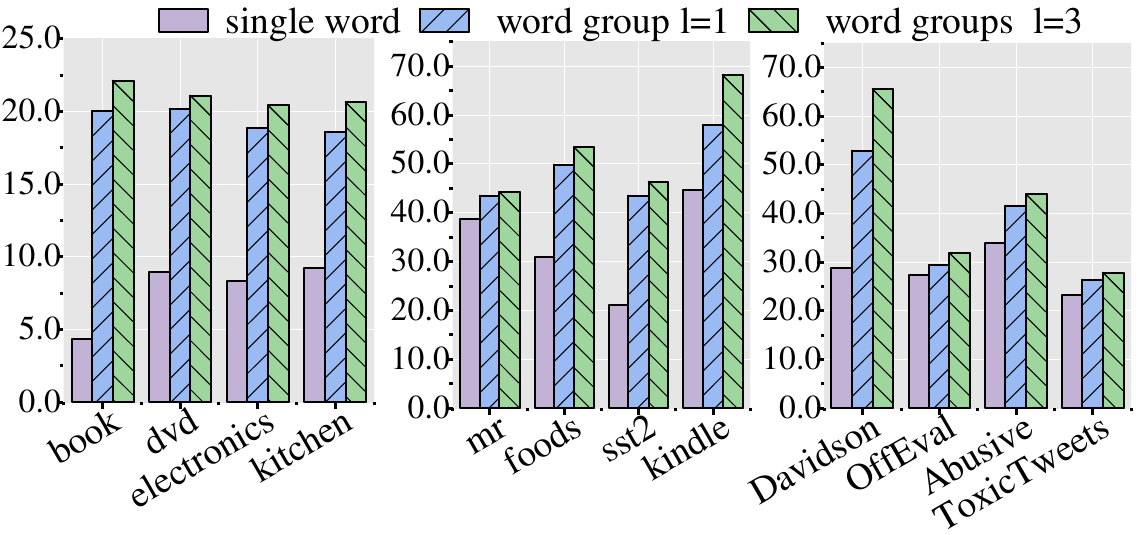}
		\label{fig:lfr}
		\caption{The label flipping rate (\%) caused by different counterfactual approaches. Single words denotes the \textbf{WO word-groups} method in Section~\ref{sec:abl}, $l=1$ denotes the \textbf{WO voting} method.}
	\end{center}
\end{figure}

The LFR scores for three cases is calculated: \textbf{single word} uses the keyword with the maximum causal effect for counterfactual substitution, \textbf{word group $l=1$} uses the word-group with the maximum causal effect for counterfactual substitution, and \textbf{word group $l=3$} denotes that if only one of the three word-groups causes a label flip, the counterfactual is successful. 

\textbf{LFR Results}. We show the corresponding results in Figure~\ref{fig:lfr}. For automatic counterfactual substitution of a single word, the values of LFR are smaller, especially for books, dvds, electronics, and kitchen that contain longer text (less than 10\%). This is due to the fact that the longer the text, the greater the number of words required for semantic flipping. When the word-groups search method is used (word group $l=1$), we can increase the LFR by searching for combinations of words of different lengths, as because word-groups contain stronger causal effect. For all datasets, $l=3$ leads to the largest LFR, because the greater the number of word-groups, the more likely it is to contain true causal features. But we also notice that $l=3$ has a small boost than $l=1$, which again shows that the number of phrases is not always better, as it introduces more noise and increases complexity. This also explains the need for the proposed voting mechanism.

Therefore, based on the above observations, we can answer \textbf{Q4}. The effectiveness of ACWG comes from the richer semantics brought by automatic counterfactual substitution, the samples after automatic counterfactual substitution are a useful agumentation to the LPLMs. The key causal features found by the word-groups search method enhance the efficiency of this counterfactual gain, thus inducing LPLMs to focus on more robust causal features.

\subsection{Q4: Case Study}\label{sec:case}

Further, we carry out in-depth analysis of the proposed framework through case analysis, so as to show the working mechanism of the model more clearly. Specifically, several sets of cases are carefully studied to explore the true effects of the proposed word-groups as shown in Table~\ref{tab:case}. For the sample from sst2, the proposed method can easily find word-group '\textit{beautiful loves}'. The word-group contains two words with a significant positive predisposition, and thus determines that the prediction is positive. For samples from foods, \textit{no purchase recommend} is found, expressing a negative assessment. Further, for the toxic cases, multiple insults are found in the sample, such as \textit{ass}, \textit{asshole}, \textit{bitch}, and \textit{pussy}. The words together constitute the toxicity of the samples, and deleting any one of them does not eliminate the toxicity of the samples. In addition, we can find that there are similarities among different word-groups of the same sample, and the voting mechanism can further strengthen the causal features by capturing such similarities. 

\section{Conclusion and Future Work}
In this paper, we propose a word-group mining method to enhance robust of large-scale pre-trained language models in the face of shortcut learning in text classification. Based on the maximum causal effect, we search the combinations of keywords to obtain robust combinations of causal features. Further, word-groups are used for automatic counterfactual generation to augment the training sample, and finally, comparative learning is used to induce model fine-tuning to improve robustness. We conduct extensive experiments on 8 sentiment classification and 4 toxic text detection datasets, and confirm that the proposed method can effectively improve the model's cross-domain generalization, robustness against attacks, and fairness.

But fine-tuning some of the existing hyperscale language models is very difficult, such as GPT-3~\cite{BrownMRSKDNSSAA20} and LLAMA~\cite{LLAMA}. Therefore, in future work, we will try to explore large-scale generative language models and analyze them from multiple perspectives for shortcut learning problems. In addition, we will study how to improve the robustness and fairness of language models by combining interpretability and prompt learning without fine-tuning.

\ifCLASSOPTIONcompsoc
  \section*{Acknowledgments}
\else
  \section*{Acknowledgment}
\fi
The authors would like to thank...

\ifCLASSOPTIONcaptionsoff
  \newpage
\fi



\bibliographystyle{IEEEtran}
\bibliography{ref}

%

\begin{IEEEbiography}[{\includegraphics[width=1in,height=1.25in,clip,keepaspectratio]{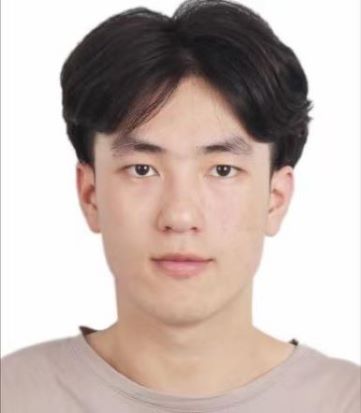}}]{Rui Song}
received a B.S. degree in the College of Software, Jilin University, China. He is currently pursuing his Ph.D. degree in the School of Artificial Intelligence, Jilin University. His research interests include fairness and robustness for large-scale pre-training language models, he also works on toxic detection of online social networks. He has published papers in IJCAI, CVPR, EMNLP, IEEE ICASSP and others. 
\end{IEEEbiography}

\begin{IEEEbiography}[{\includegraphics[width=1in,height=1.25in,clip,keepaspectratio]{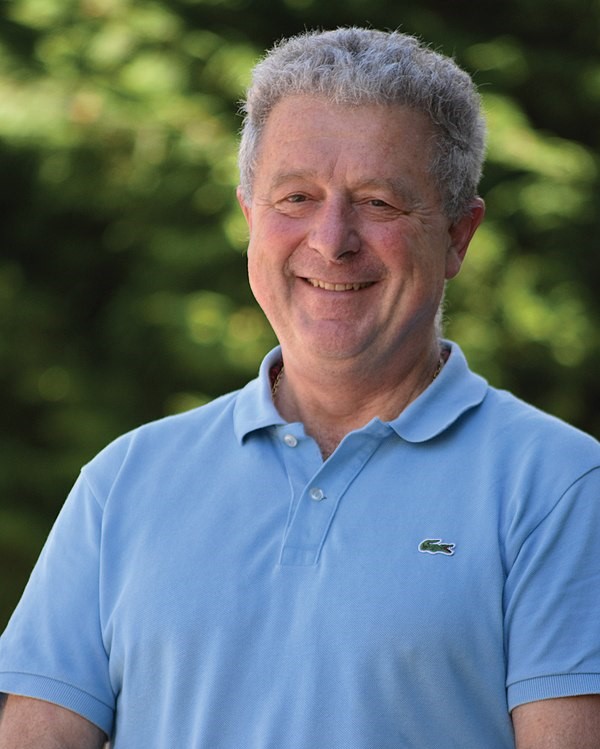}}]{Fausto Giunchiglia}
	is a member of the European Academy of Sciences, professor at the University of Trento, Italy, and Chief Scientist at the International Center of Future Science of Jilin University. He was Vice President of the University of Trento, Italy, founder and head of the Department of Computer Science, President of the IJCAI-2005 and President of the WWW-2021. His research interests focus on artificial intelligence, formal methods, knowledge management and agent-oriented software engineering. 
\end{IEEEbiography}

\begin{IEEEbiography}[{\includegraphics[width=1in,height=1.25in,clip,keepaspectratio]{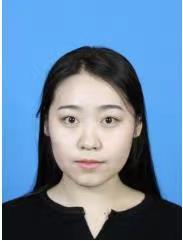}}]{Yingji Li}
	received her bachelor's degree from the College of Computer Science and Technology at Jilin University in 2019.  She is currently a Ph.D. student in the College of Computer Science and Technology at Jilin University. Her research interests include natural language processing, graph representation learning, and fairness in deep learning. She has published papers in ACL and others.
\end{IEEEbiography}

\begin{IEEEbiography}[{\includegraphics[width=1in,height=1.25in,clip,keepaspectratio]{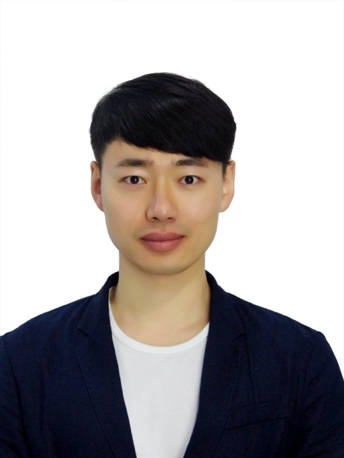}}]{Hao Xu} 
	got his Ph.D. at the University of Trento, Trento, Italy, and he is a Professor at College of Computer Science and Technology, Jilin University, Changchun, China. His research interests include artificial intelligence, digital humanities, and human-computer interaction. He has published papers in CVPR, IJCAI, ACM MM, ACL, EMNLP, IEEE ICASSP, IEEE TGRS and others.
\end{IEEEbiography}




\end{document}